\pdfoutput=1

\documentclass[11pt]{article}

\usepackage{EMNLP2023}

\usepackage{times}
\usepackage{array} 
\usepackage{latexsym}

\usepackage[T1]{fontenc}

\usepackage[utf8]{inputenc}

\usepackage{times}
\usepackage{latexsym}
\usepackage{xtab} 
\usepackage{tabularx}
\usepackage{graphicx}
\usepackage{amsmath,amssymb}
\usepackage{subfig}
\usepackage{stmaryrd}
\usepackage{multirow}
\usepackage{longtable}
\usepackage{tabularray}
\usepackage{booktabs}
\usepackage{subcaption}
\usepackage{colortbl}
\usepackage{caption}
\usepackage{titlesec}
\usepackage{booktabs}
\usepackage{enumitem}

\newif\ifcomments
\commentstrue
\ifcomments
    \providecommand{\jb}[1]{{\protect\color{blue}{[JB: #1]}}}
    \providecommand{\sa}[1]{{\protect\color{red}{[SA: #1]}}} 
    \providecommand{\am}[1]{{\protect\color{magenta}{[AM: #1]}}} 
\else
    \providecommand{\jb}[1]{}
    \providecommand{\sa}[1]{}    
    \providecommand{\am}[1]{}    
\fi

\usepackage{inconsolata}

\title{\emph{When the LM misunderstood the human chuckled}: \\
Analyzing garden path effects in humans and language models
}

\begin{document}

\author{Samuel Joseph Amouyal$^{\lambda}$ \hspace{0.5cm} Aya Meltzer-Asscher$^{\dagger,\star}$ \hspace{0.5cm} Jonathan Berant$^{\lambda}$ \\ 
$\lambda$ Blavatnik School of Computer Science, Tel Aviv University, Israel \\
$\dagger$ Department of Linguistics, Tel Aviv University, Israel \\
$\star$ Sagol School of Neuroscience, Tel Aviv University, Israel \\
\texttt{\{samuel.amouyal, joberant\}.cs.tau.ac.il} \\
\texttt{ameltzer@tauex.tau.ac.il}} 

\maketitle

\begin{abstract}

Modern Large Language Models (LLMs) have shown human-like abilities in many language tasks, sparking interest in comparing LLMs' and humans' language processing. In this paper, we conduct a detailed comparison of the two on a sentence comprehension task using garden-path constructions, which are notoriously challenging for humans. Based on psycholinguistic research, we formulate hypotheses on why garden-path sentences are hard, and test these hypotheses on human participants and a large suite of LLMs using comprehension questions. Our findings reveal that both LLMs and humans struggle with specific syntactic complexities, with some models showing high correlation with human comprehension.
To complement our findings, we test LLM comprehension of garden-path constructions with \emph{paraphrasing} and \emph{text-to-image generation} tasks, and find that the results mirror the sentence comprehension question results, further validating our findings on LLM understanding of these constructions.
\end{abstract}
\section{Introduction}
Large Language Models (LLMs) have shown high proficiency in language comprehension and generation, demonstrating performance that matches and sometimes surpasses human capabilities across a range of tasks \cite{gpt4, llama2023, falcon40b, llama3herdmodels, geminiteam2024gemini}. This has sparked a line of research focused on comparing sentence-processing mechanisms in humans and LLMs. Within this research, some studies have found correlations between LLM activations and brain activations during the processing of identical sentences \cite{cacheteux-middle-layer, fedorenko_brain_corr, Ren2024DoLL}. Others have demonstrated that LLMs can be used to predict human linguistic behavior \cite{linzen-etal-2016-assessing, acceptability_nn, hu-etal-2020-systematic, Rego2024LanguageMO, Sun2024ComputationalSM, kuribayashi2025large}.

\begin{figure}[t!]
    \centering
    \includegraphics[width=8.0cm, height=8cm, trim={0.8cm 0.5cm 0cm 0.4cm},clip]{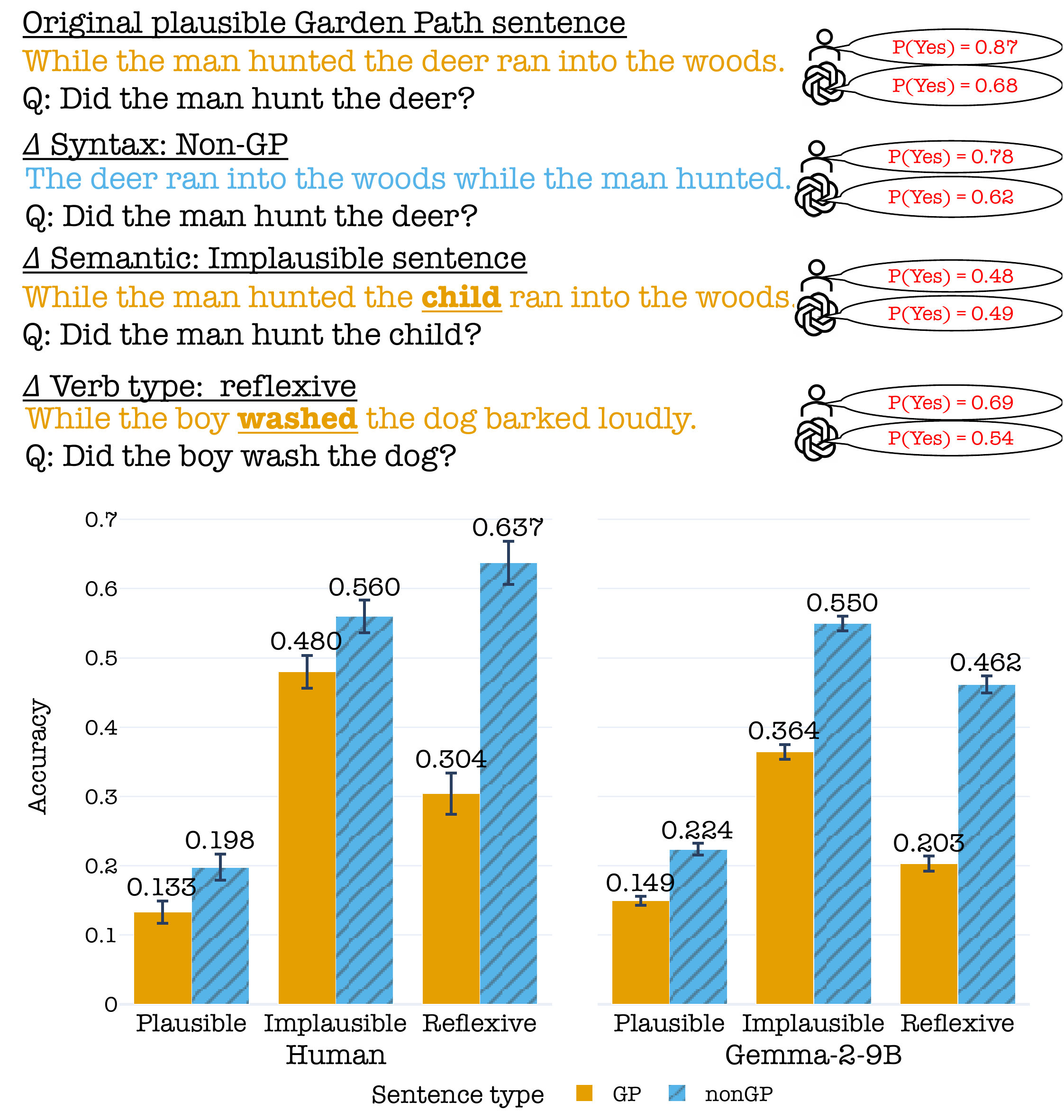} 
    \caption{\emph{Top}: The manipulations made to an example garden-path sentence along with predictions from humans and LLMs for these sentences. \emph{Bottom}: human and the Gemma-2-9B average performance on the different experimental conditions. The behaviour of humans and Gemma-2-9B is similar.}
    \label{fig:human_next_llm}
\end{figure}

\begin{figure*}[t]
    \centering
    \scriptsize
    \subfloat[\centering While the boy washed the dog barked loudly.]{{\includegraphics[width=6cm, height=5cm]{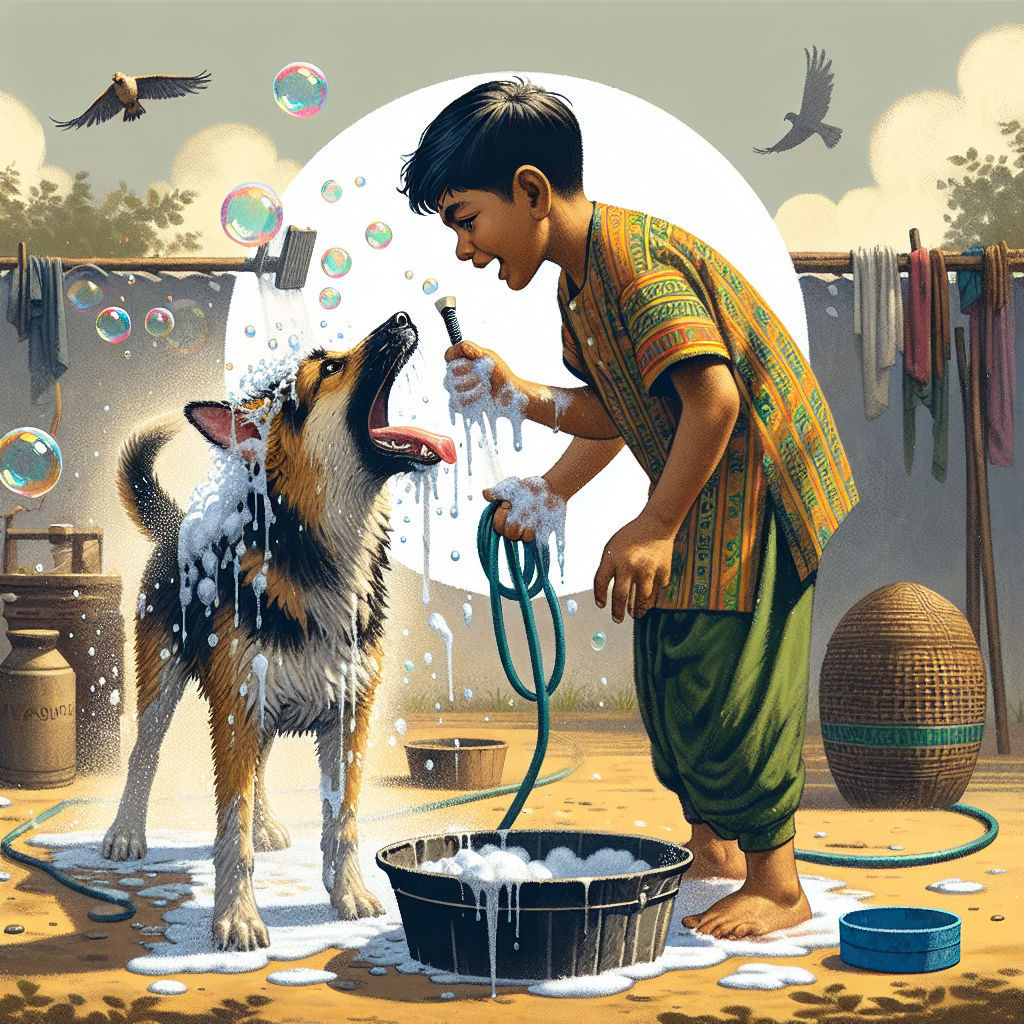}}}%
    \qquad
    \subfloat[\centering The dog barked loudly while the boy washed.]{{\includegraphics[width=6cm, height=5cm]{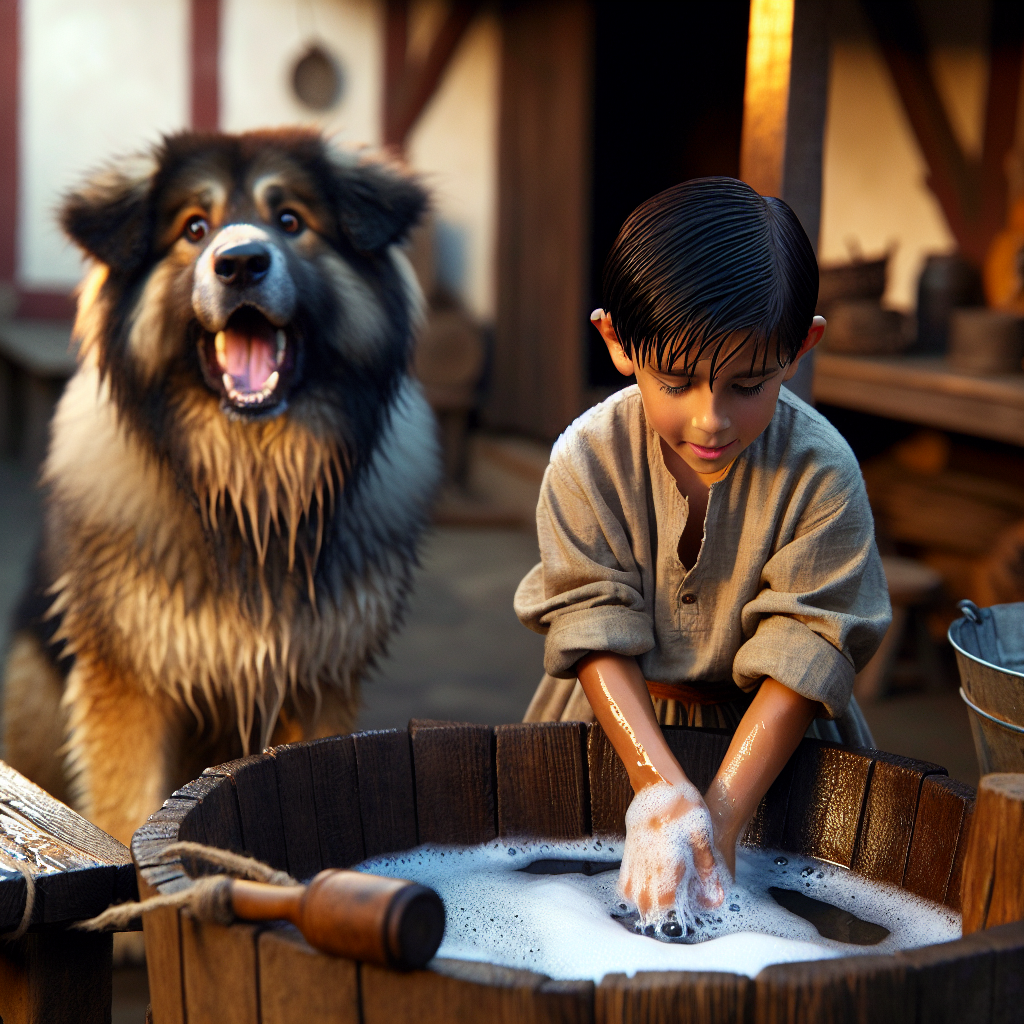}}}
    \caption{Dall-e-3 incorrectly generates an image where the boy washes the dog (Left) given a GP sentence, but generates a correct image with a non-GP sentence (Right).}
    \label{fig:dalle_fig1}
\end{figure*}

While LLMs mostly succeed where humans succeed, less is known on whether LLMs fail where humans fail. A classic case in psycholinguistic research for sentences where humans systematically have comprehension difficulty is Garden Path (GP) structures \cite{gp3, gp2, gp1}. GP sentences are temporarily ambiguous, as their beginning leads readers to misconstrue their parse. Consider for example (\ref{num:ex_gp})-(\ref{num:ex_ngp}). In (\ref{num:ex_gp}), readers initially misanalyze ``the dog'' as the object of ``washed'', although in the final, correct structure, ``the dog'' is the subject of ``barked'', and ``washed'' has no object. 
\begin{enumerate}[nosep]
    \item \label{num:ex_gp} While the boy washed the dog barked loudly. 
    \item \label{num:ex_ngp} The dog barked loudly while the boy washed.
\end{enumerate}
\citet{christianson2001} used comprehension questions and found that comprehension is poorer for GP sentences like (\ref{num:ex_gp}) compared to non-GP sentences like (\ref{num:ex_ngp}). Specifically, the initial interpretation, where the post-verbal NP is analyzed as the object of the first verb, lingers, while the NP is also analyzed as the subject of the following verb (see also \citet{christianson2006}). \citet{patson2009lingering} similarly showed that when participants were tasked with paraphrasing GP sentences,  they often gave paraphrases such as ``The boy washed the dog and the dog barked'' for (\ref{num:ex_gp}). 

Some prior studies examined how LLMs handle GP sentences. \citet{gp_reading_time} found that LLM-based metrics significantly underestimate the processing difficulties humans face with GP sentences. \citet{hanna2024incrementalsentenceprocessingmechanisms} found that LMs encode multiple parses of GP sentences, which does not straightforwardly align with human performance. \citet{bert_gp} reported that BERT misinterprets GP sentences, although with different error patterns than humans. In contrast, \citet{Li2024IncrementalCO} observed that four LLMs make parsing errors akin to human errors. Despite these findings, gaps remain in existing research. First, it remains unclear whether the same aspects of GP structures cause processing difficulties in humans and LLMs. Second, the breadth of LLM families considered has been limited thus far.
Lastly, the comparison between humans and LLMs has been mostly carried out through indirect measures, where e.g. human reading times are correlated with LLM uncertainty \cite{wilcox-gpt2-abilities, Rego2024LanguageMO, Sun2024ComputationalSM}, whereas a comparison on the same task could be more revealing.


\def\thickhline{\noalign{\hrule height.8pt}}
\newcolumntype{?}{!{\vrule width .8pt}}

In this study, we explore object/subject GP sentences (similar to (\ref{num:ex_gp})) with humans and LLMs, with both responding to exactly the same task, namely, a comprehension question about the sentence (e.g., ``Did the boy wash the dog?'').
We present three (non-mutually exclusive) hypotheses regarding the challenges posed by GP sentences: (a) Misinterpretation of the noun phrase as the object of the verb arises due to the difficulty of syntactic reanalysis; (b) Misinterpretation arises since the noun phrase is a plausible object for the verb (e.g., boys tend to wash dogs); (c) Misinterpretation arises since transitive verbs (\emph{``hunt''}) entail some object. Reflexive (\emph{``wash''}) or unaccusative (\emph{``drop''}) verbs, which are interpreted with no direct object, will give rise to less misinterpretation.


\begin{table*}[t!]
    \scriptsize 
    \centering
        \begin{tabular}{ l ? l}
         \textbf{Hypothesis} & \textbf{Manipulation} \\
        \thickhline
        GP syntax is hard& While the man hunted the deer ran into the woods. $\rightarrow$ The deer ran into the woods while the man hunted. \\
        \hline
        Plausible direct object& While the man hunted the deer ran into the woods. $\rightarrow$ While the man hunted the child ran into the woods. \\
        \hline
        Transitive vs. reflexive/unaccusative& While the man hunted the deer ran into the woods. $\rightarrow$ While the boy washed the dog barked loudly. \\
        \end{tabular}
        \vspace{-0.2cm}
    \caption{We manipulate the GP structures examined to test three hypotheses for what makes GP sentences hard. }
    \label{tab:changes}
\end{table*}


We test our hypotheses on sets of sentences instantiating the various manipulations (see Table~\ref{tab:changes}) and tested comprehension in humans and LLMs.



Our human results (Fig. \ref{fig:human_next_llm}, left) provide evidence for all three hypotheses. Accuracy is lower (i) when reanalysis is needed (i.e. in GP compared to non-GP structures), (ii) when the noun is a plausible (compared to implausible) object for the verb, and (iii) when the verb is transitive, entailing an object, compared to reflexive/unaccusative.  The second, semantic effect is stronger than the syntactic and verb type effects. 

Interestingly, we find that LLMs struggle with comprehension of GP sentences, even in the most capable models (o1-preview accuracy is highest at 78\%). In addition, for many LLMs, the manipulations have similar effects to those they have on humans. Examining multiple models within the same families reveals that stronger models tend to display greater similarity to human behavior, as shown in Figure \ref{fig:human_next_llm} (right) for Gemma-2-9B.


Finally, we validate our results for LLMs with two additional sentence comprehension tasks -- paraphrasing and image generation. Both tasks produced results akin to the comprehension questions task. Figure~\ref{fig:dalle_fig1} shows Dall-e-3's \cite{dalle_3} inability to correctly parse the GP sentence (depicting the boy washing the dog instead of himself), unlike in the non-GP sentence.


To summarize, our contributions are:
\begin{enumerate}[nosep]
    \item We put forward precise hypotheses for the challenges that GP sentences pose and construct linguistic materials to test them.
    \item We collect human and LLM data on a \emph{sentence comprehension} task to test these hypotheses .
    \item We compare LLM performance to humans, finding that stronger models are more similar to humans. 
    \item We validate our results on a paraphrasing and text-to-image task.
\end{enumerate}

\section{What Makes Object/Subject Garden-Path Sentences Hard?}

Object/subject garden-path sentences, like (\ref{num:ex_gp}), include an embedded verb (``wash'') followed by a noun phrase (``the dog'') and the main verb (``barked''). These sentences cause processing difficulties, leading to slower reading—especially at the main verb—and reduced comprehension. In sentence (\ref{num:ex_gp}), \emph{``the dog''} will often be mistakenly interpreted as the object of \emph{``wash''}, prompting incorrect answers like \emph{``Yes''} to \emph{``Did the child wash the dog?''}. This happens even though in the final structure of the sentence, ``the dog'' is not an object of ``washed''. 


\paragraph{}
Several (non-mutually exclusive) hypotheses can explain the misinterpretation described above, as summarized in Table~\ref{tab:changes}.
To describe the hypotheses, we consider the following sentence-question pair:
\begin{enumerate}[nosep,leftmargin=*]
    \setcounter{enumi}{2}
    \item \label{num:ex_gp_3} While the man hunted the deer ran into the woods.
\end{enumerate}
\begin{itemize}[nosep,leftmargin=*]
    \item[] \textbf{Question}: Did the man hunt the deer?   Y/N
\end{itemize}

Note that the accurate answer to the question above is ``Not necessarily''. It is a possible interpretation of the sentence, and may be inferred from it, but it is not entailed from the sentence. In our experiments, as in previous experiments, we consider "yes" to be a wrong answer here, whereas "no" is considered the right answer.  

\textbf{Hypothesis 1: The GP syntax is hard.}
    This hypothesis suggests that misinterpretation occurs because during incremental processing, the post-verbal noun phrase \emph{(``the deer'')} is first attached as the object of the verb, requiring reanalysis when the second verb is encountered. Often, the reanalysis is not complete, and the initial interpretation lingers. According to this, reordering the clauses (see Table~\ref{tab:changes}) should improve accuracy by preventing initial misattachment.

 \textbf{Hypothesis 2: Readers attach the noun to the first verb when it is a plausible object for it. } 
    According to this hypothesis, readers interpret a noun as an object of a verb in the sentence whenever this is semantically plausible, regardless of sentence position. If the noun is an implausible direct object, it will not be interpreted as such, improving accuracy (see Table~\ref{tab:changes}).\textbf{
    Hypothesis 3: Readers search maximal interpretation of verb arguments.} 
    According to this hypothesis, optionally transitive verbs need objects for full interpretation, so available nouns are taken to fulfill this role. In contrast, alternating reflexive (``wash'') and unaccusative verbs (``drop'') allow complete interpretation intransitively, without an object (e.g., \emph{The boy washed}, \textit{The ball dropped}), Thus, such verb will allow more accurate interpretation than optionally transitive verbs (see Table~\ref{tab:changes}).
    Note that for sentences such as "While the boy washed the dog barked", the correct answer for ``Did the boy wash the dog?'' is ``No''. 
    As for the optionally transitive verbs, it can also be hypothesized that the tendency to interpret them as taking an object depends on the verb's transitivity bias (the probability that the verb appears with a direct object). According to this, verbs with a lower bias (e.g., ``walk'') should lead to better accuracy compared to those with a higher bias (e.g., ``explore''), as the noun is less likely to be considered as their object.

Next, we describe how we test the above hypotheses, starting with a human experiment.
\section{Human performance}

We first run an experiment on human participants to test our hypotheses.

\subsection{Methods}

\paragraph{Materials} 
In 45 sentences sets with optionally transitive verbs (24 coming from \citet{christianson2001}, 21 crafted for this study), we manipulated the structure of the sentence (GP or non-GP) and plausibility of the noun as the verb's object (plausible or implausible), as exemplified in (\ref{num:ex_4a}). We also created 24 additional sets with reflexive/unaccusative verbs (12 from \citet{christianson2001} and 12 crafted for this study) in plausible sentences, manipulating structure (GP/non-GP), as exemplified in (\ref{num:ex_5a}). 

\begin{enumerate}[nosep]
    \setcounter{enumi}{3}
    \small \item \label{num:ex_4a} \begin{enumerate}[nosep]
        \item GP, plausible: While the man hunted the deer ran into the woods.
        \item Non-GP, plausible: The deer ran into the woods while the man hunted. 
        \item GP, implausible: While the man hunted the child ran into the woods.
        \item Non-GP, implausible: The child ran into the woods while the man hunted.
        \end{enumerate}
        \item \label{num:ex_5a} \begin{enumerate}[nosep]
            \item  GP, reflexive: While the boy washed the dog barked loudly.
            \item Non-GP, reflexive: The dog barked loudly while the boy washed.
        \end{enumerate}
    \end{enumerate}

To construct materials for the plausibility manipulation  (hypothesis 2), we use
insights from \citet{amouyal-etal-2024-large}, and let GPT4 rate sentence plausibility on a 1 to 7 scale. We select pairs where the plausible sentence had a rating of at least 3 points higher than its implausible counterpart. For the second part of hypothesis 3, we assessed each optionally-transitive verb's bias by its proportion of transitive usages on Wikipedia. Our verbs' bias ranges from 0.102 (``sail'') to 0.775 (``explore''). Appendix \ref{sec:verb_factors} lists the full estimated verb biases, and all the sentences are in Appendix \ref{sec:appendix_sents}.


For each sentence, we ask one of two questions:
\begin{enumerate}[nosep]
    \item \emph{Simple}: \emph{``Did the deer run into the woods?''}
    \item \emph{GP}: \emph{``Did the man hunt the deer?''}
\end{enumerate}
The simple question probes basic understanding of the sentence, whereas the GP question targets the potential misinterpretation. 

\begin{figure*}[t!]
    \centering
    \scriptsize
    \includegraphics[width=17cm]{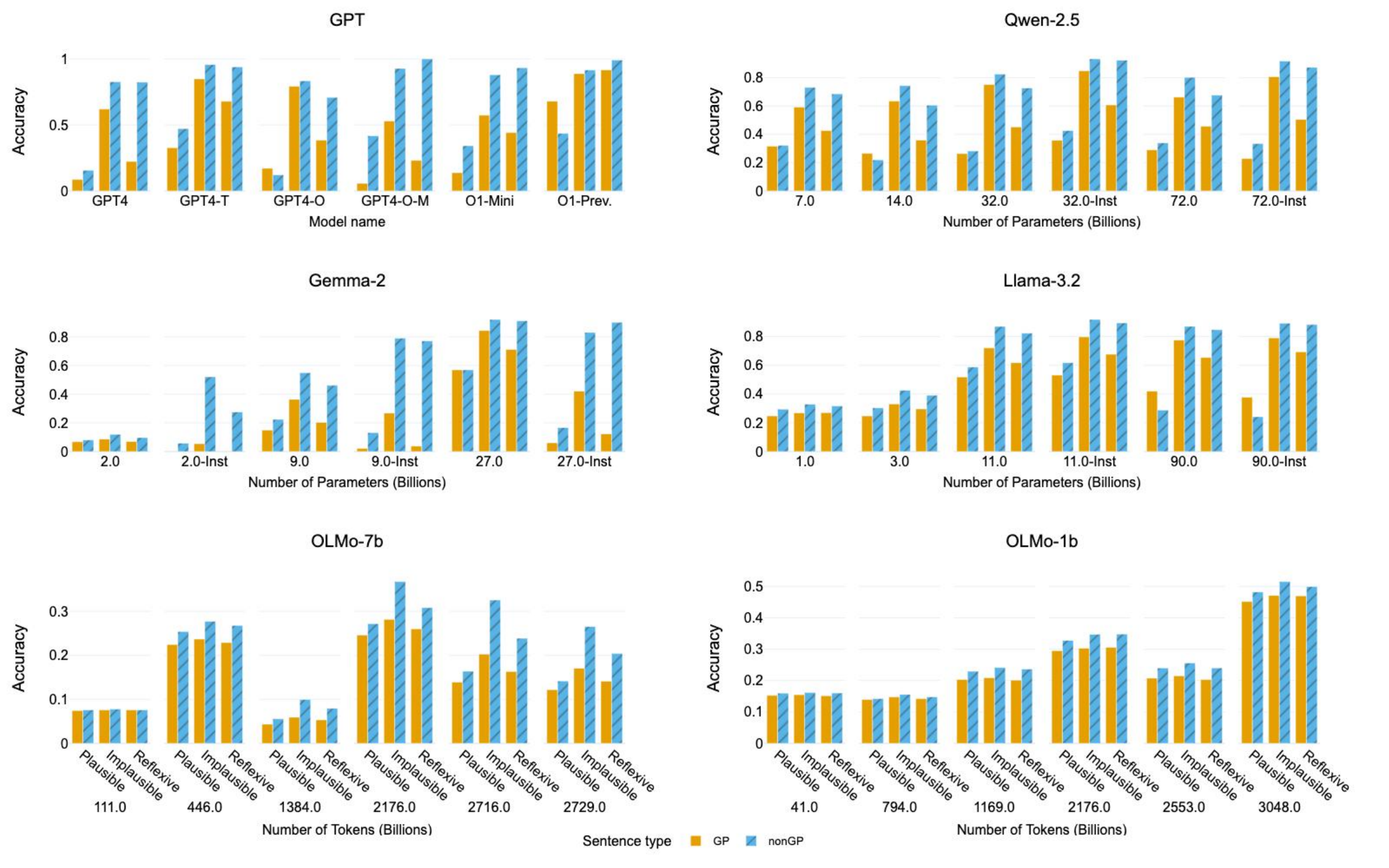}
    \caption{Performance of models from all families on our experimental conditions. Models with an ``-Inst'' suffix are instruction-tuned. Note: Each figure has a different y-axis range.}
    \label{fig:model_global_perf}
\end{figure*}

\paragraph{Procedure} 

Native English speakers were recruited via the Prolific platform.\footnote{\url{https://www.prolific.com/}}  Sentences were displayed word-by-word, with each word shown for 400ms and a 100ms blank screen between words. After the sentence, the comprehension question was presented for 5 seconds. If unanswered within this time, the response was marked as incorrect. Participants completed two practice items, followed by one experimental sentence and one question. The single-trial design prevents fatigue \cite{christianson2022if} and learning effects \cite{fine2013rapid}. Each of the 456 sentence-question pair was shown to 10 participants. The average completion time was 1:50 minutes, and participants were compensated with 0.30£, equivalent to 9.64£ per hour. 
The experiment was approved by the Ethics Committee at Tel-Aviv University. 


\subsection{Human results}

The accuracy on simple questions was high (average 95.4\%, minimum 92.4\%, maximum 98.7\%).
Conversely, GP questions were much more challenging with an average accuracy of 37.0\% on GP questions.

Figure \ref{fig:human_next_llm} (left) shows the average accuracy for humans in the various conditions. As expected from Hypothesis 1, accuracy is consistently higher for non-GP structures. In addition, accuracy is lower when the noun is a plausible direct object, indicating a tendency to interpret it as such even without syntactic indication, supporting Hypothesis 2. The plausibility effect was more pronounced than the syntactic effect. In addition, accuracy was higher for GP sentences with a reflexive/unaccusative verb than for those with an optionally transitive verb, supporting Hypothesis 3, and the effect of structure (GP vs. non-GP was stronger for the former verbs). 


We test statistical significance with Generalized Linear Mixed-Effects Models (see Appendix \ref{sec:glmm_app}). For Hypothesis 1, the difference between GP and non-GP sentences was significant for implausible ($p$ = .019) and reflexive ($p$ = 2.52e-13) sets, and approached significance in the plausible sets, $p$ = .065. Hypothesis 2's prediction was confirmed with a significant difference between plausible and implausible sentences ($p$ = 4.11e-16). Hypothesis 3 is also supported with a significant difference between reflexive/unaccusative verbs for both GP sentences ($p$ = 1.35e-5) and non-GP sentences ($p$ = 2.45e-14). The Pearson correlation between transitivity bias and accuracy was weak ($\leq$0.19) across conditions. 


Overall, the hypotheses were supported by the human results, showing that multiple factors influence the difficulty of object/subject GP sentences.

\section{LLMs Performance}

We now analyze the performance of LLMs on our different experimental conditions. 

\subsection{Methodology}

To replicate the experiment with LLMs, we used few-shot prompting, where each example includes a sentence, a question, and the correct answer.\footnote{We also experimented with chain-of-thought-prompting, which did not lead to a signifnicant change in results.} The examples did not contain GP structures. Each model was prompted 8 times, using two system prompts and four example orderings. We extract the probabilities of the correct and incorrect answers tokens, averaging these across the 8 prompts. Appendix \ref{sec:prompt_example} shows an example prompt.


\paragraph{Models}

We test models from different families, sizes and training checkpoints:
\begin{enumerate}[nosep]
    \item GPT family \cite{gpt4}: \emph{GPT-4}, \emph{GPT-4-Turbo}, \emph{GPT-4o}, \emph{GPT-4o-mini}, \emph{o1-preview}, \emph{o1-mini}.
    \item Llama-3 \cite{llama3herdmodels}: All models from the Llama-3 family (Llama-3.2 and Llama-3.1) available on HuggingFace.\footnote{\url{https://huggingface.co/models}}
    \item Qwen-2.5 \cite{qwen2, qwen2.5}: All Qwen-2.5 models on HuggingFace except models of size 0.5b.
    \item Gemma-2 \cite{gemma2}: All Gemma-2 models on HuggingFace.
    \item Olmo \cite{groeneveld2024olmoacceleratingsciencelanguage}: 15 Olmo-1b and Olmo-7b checkpoints along training.
\end{enumerate}

\subsection{Results}

We first present the overall results of LLMs on our task.  Figure \ref{fig:model_global_perf} shows the results for 6 selected models from each family. Appendix \ref{sec:all_model_res} presents the results for all models.

At a high-level it is clear that the behavior of LLMs resembles that of humans: accuracy on non-GP sentences is higher than accuracy on GP sentences, accuracy for both GP and non-GP sentences is higher when the direct object is implausible, and the gap between GP and non-GP sentences is larger in the reflexive/unaccusative case. These trends seem more pronounced for larger and stronger models (the top two rows) compared to smaller models (Olmo-1b).

Interestingly, LLM performance is not perfect even for the strongest model, \emph{o1-preview}, which obtains an average accuracy of 78\% (the second-strongest model, \emph{Gemma-27B} has an average accuracy of 74\%). 
This far-from-perfect performance of LLMs is perhaps surprising since the entire sentence and question are presented in full to the LLMs and there is no reason to  suspect that they should suffer from the same processing difficulties that humans do, especially those related to the inability to overcome the initial misparse.


\begin{figure}[t!]
    \centering
    \scriptsize
    \includegraphics[height=8cm]{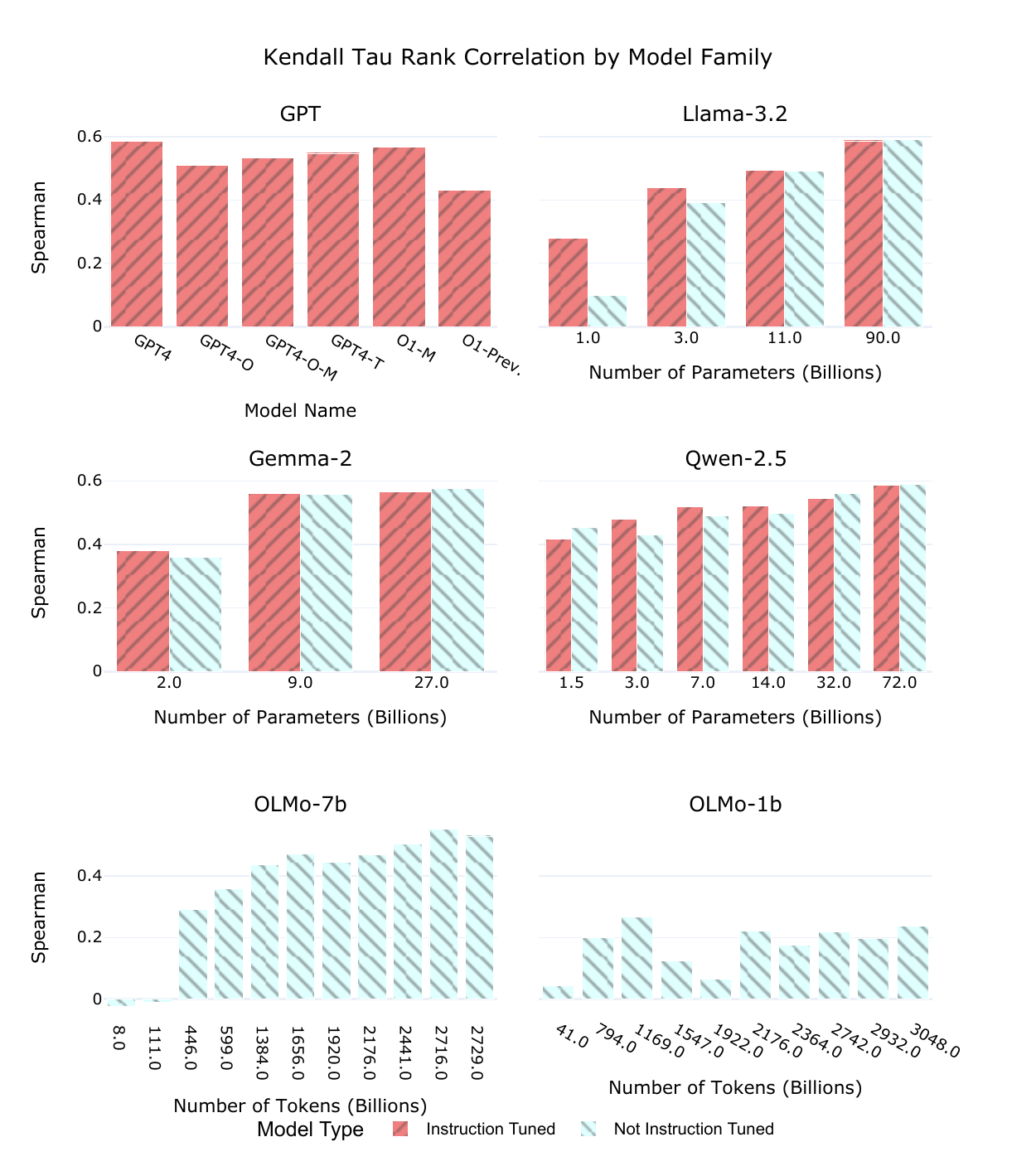}
    \caption{Gloabal Kendall Tau rank correlation per model family.}
    \label{fig:kendall_tau_global}
\end{figure}


\section{Analzying LLMs vs. human performance}

In comparing LLMs and humans, we focus on the following important (albeit less-studied) aspect: the extent to which the relative difficulty of tasks in our experiment is similar between LLMs and humans. Showing that LLMs and humans have similar processing difficulties can open interesting research directions on whether LLMs can inform psycholinguistic models of human sentence processing \cite{kuribayashi2025large}.

To evaluate the similarity between LLMs and humans, we use the Kendall Tau rank correlation metric.\footnote{\url{https://en.wikipedia.org/wiki/Kendall_rank_correlation_coefficient}. Our human data has a lot of ranked ties, and Kendall Tau correlation accounts for ranked ties contrary to Spearman rank correlation.} We calculate a \emph{``Global''} Kendall Tau correlation by looking at all items in our data (from all conditions) and comparing the average accuracy of humans on those items to the average probability of the correct answer as provided by the LLM, 
This measures whether the difficulty ranking of all the items on our experiment is similar for humans and LLMs. Figure \ref{fig:kendall_tau_global} presents the findings for each model family.

\paragraph{Model size:} As illustrated in Figure \ref{fig:kendall_tau_global}, larger models exhibit a higher Kendall Tau correlation with human judgements across all model families.

\begin{figure}[t!]
    \centering
    \scriptsize
    \includegraphics[height=8cm]{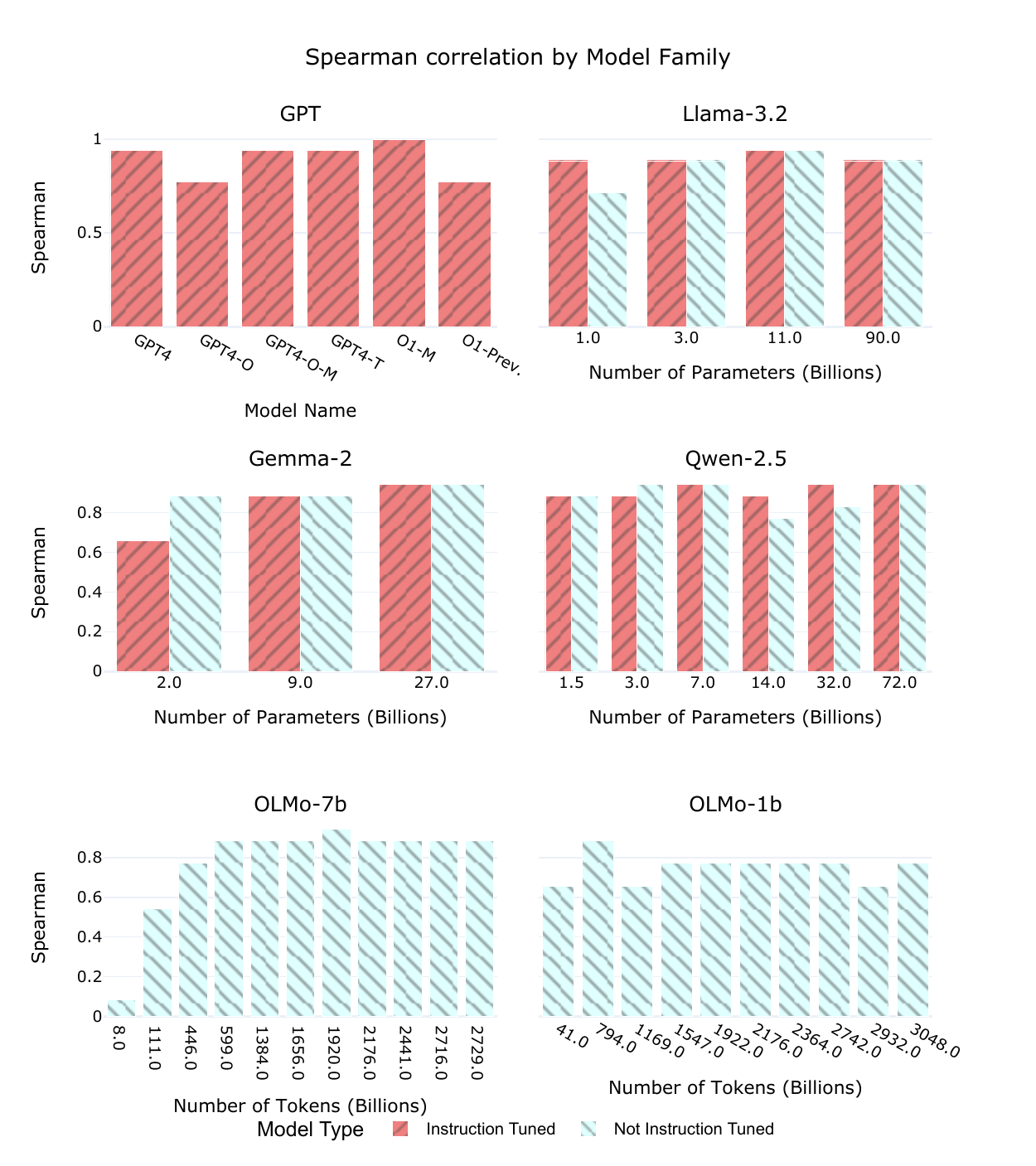}
    \caption{Spearman rank correlation per model family}
    \label{fig:spearman_per_category}
\end{figure}

\begin{figure*}[t!]
    \centering
    \scriptsize
    \includegraphics[width=18cm]{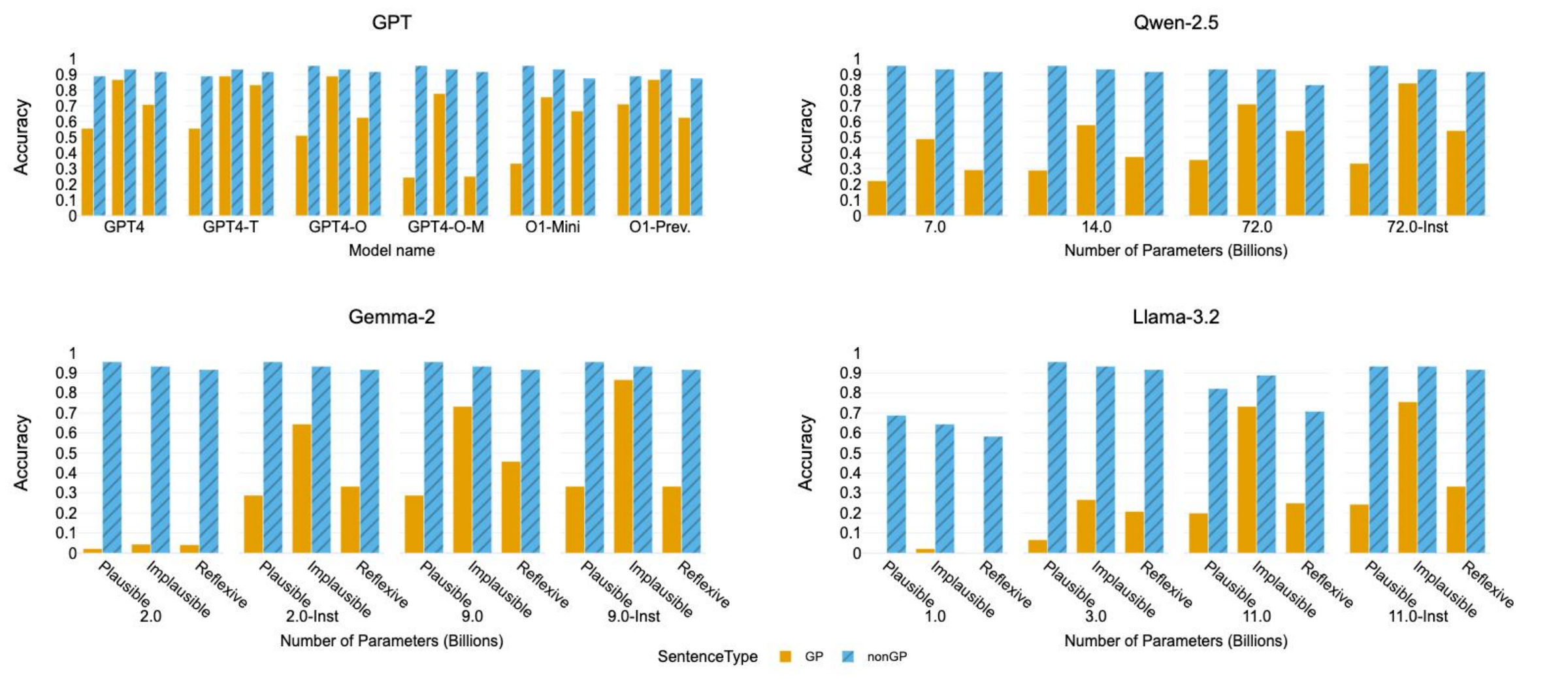}
    \caption{Average paraphrase accuracy for each condition per model family}
    \label{fig:paraphrase_acc}
\end{figure*}

\paragraph{Instruction tuning:} Instruction tuning appears to have little impact on the similarity between humans and LLMs in this measure.

\paragraph{Pretraining tokens:} In OLMo-7B, we see an increase in Kendall Tau correlation with the increase in number of pretraining tokens. This pattern is not observed in OLMo-1B, possibly because the model is too weak to show significant effects. 

We now move to analyzing the by-condition correlation between humans and LLMs.
Figure \ref{fig:spearman_per_category} shows the Spearman rank correlation between humans and LLMs, comparing the \emph{average accuracy} across the 6 conditions (correlating two vectors in $\mathbb{R}^6$). All models show a high Spearman rank correlation with human data. This suggests that models align well with humans in ranking average accuracy by sentence type, as opposed to a global ranking across all items. Notably, model size minimally impacts Spearman rank correlation but significantly affects Kendall Tau, indicating that larger models better differentiate item difficulty within conditions. In OLMo-7b, as training progresses, condition difficulty distinctions increasingly resemble humans'.



\section{Performance on Paraphrasing and Text-to-Image Generation}

\begin{figure*}[t]
    \centering
    \scriptsize
    \subfloat[\centering The yacht headed toward a waterfall while the explorer paddled.]{{\includegraphics[width=3cm]{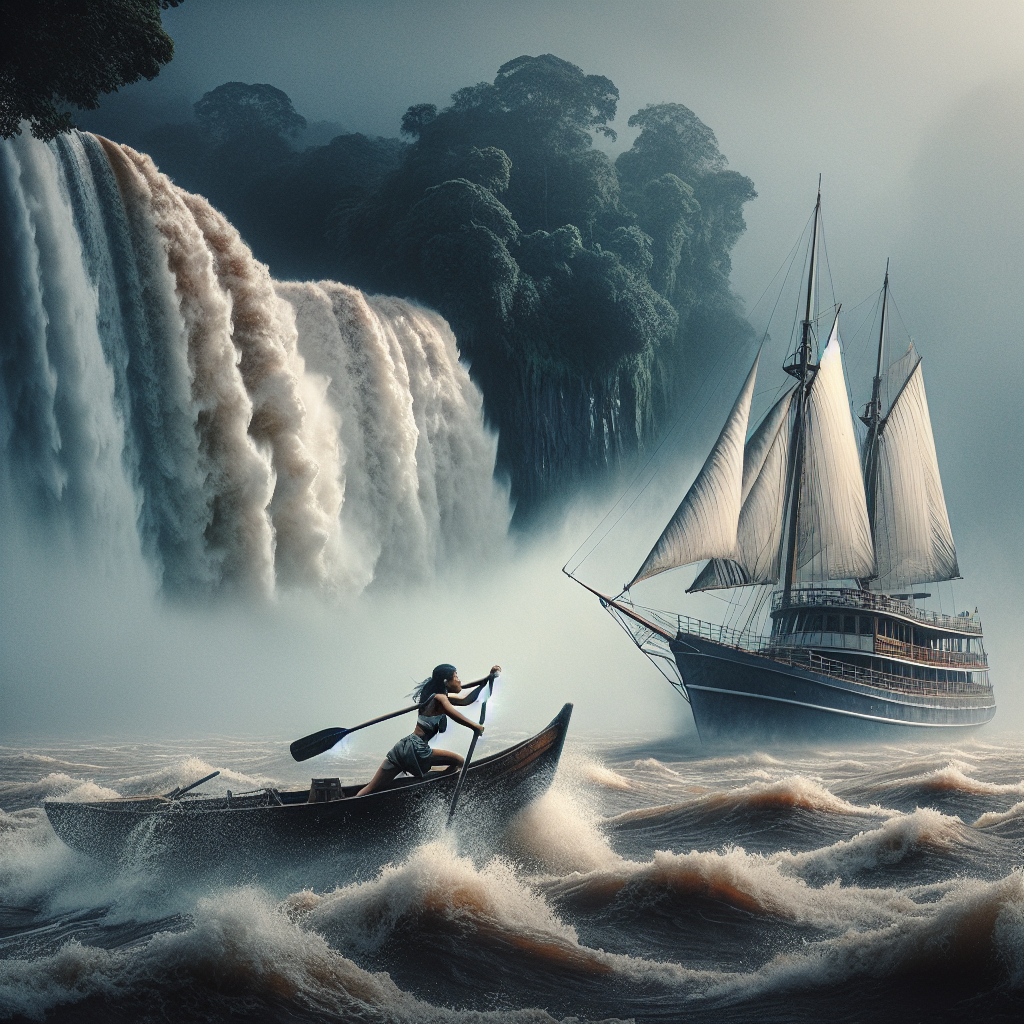}}}%
    \qquad
    \subfloat[\centering The rainbow slowly faded outside while the girl painted.]{{\includegraphics[width=3cm]{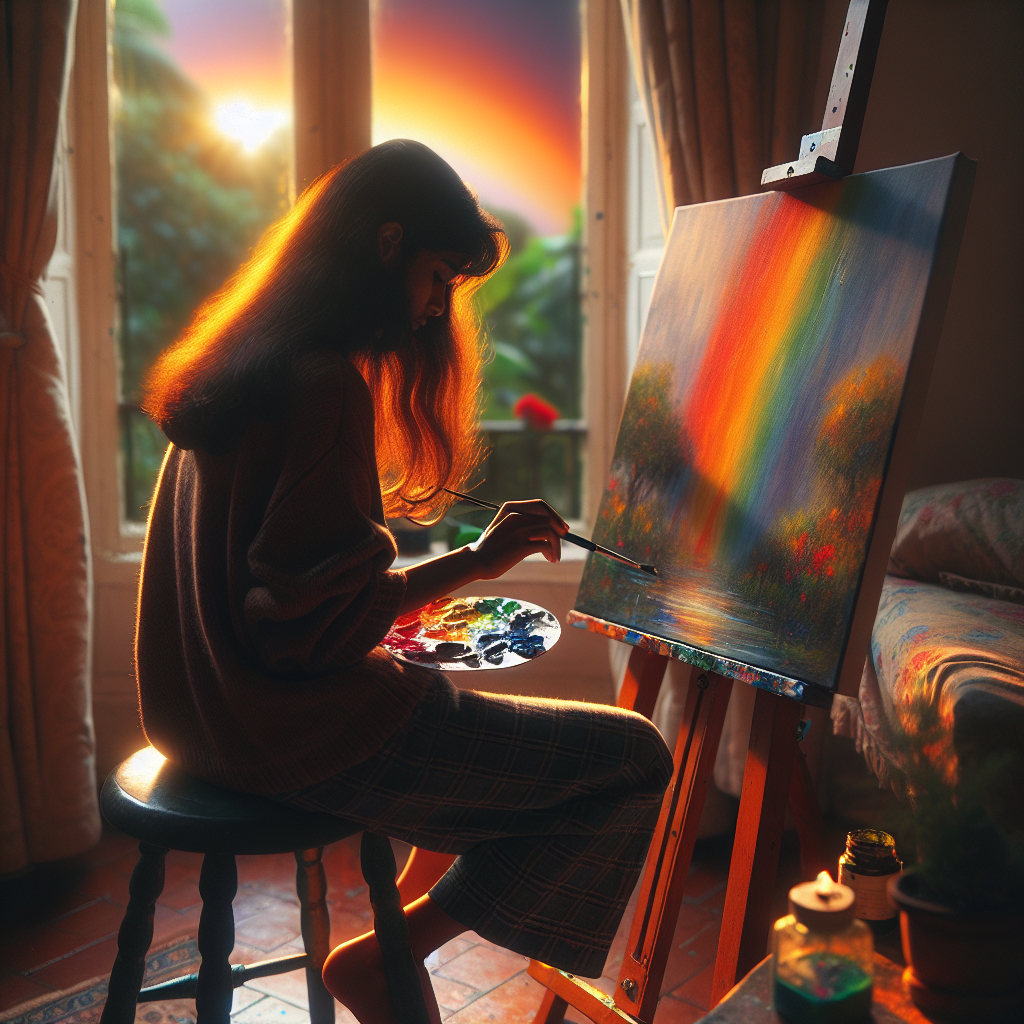}}}
    \qquad
    \subfloat[\centering While the teacher counted the children formed a line.]{{\includegraphics[width=3cm]{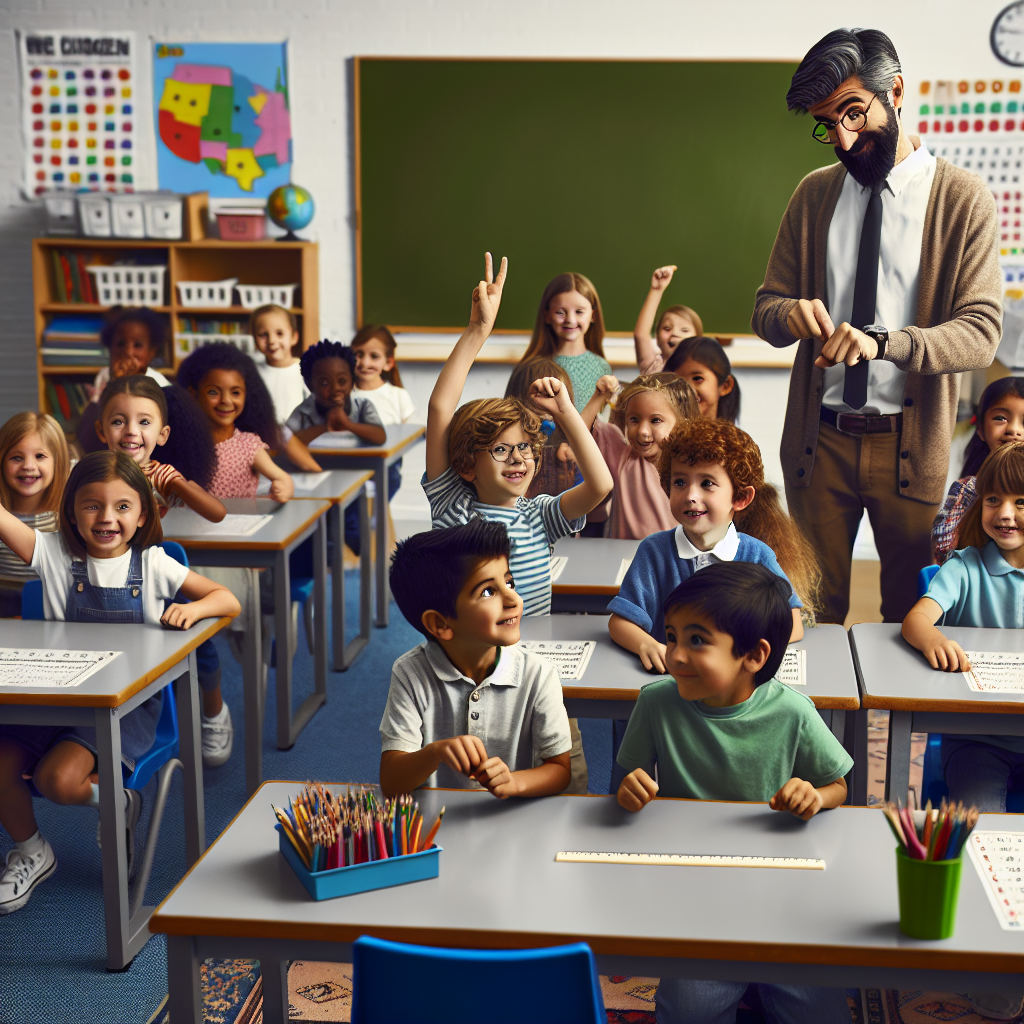} }}
    \qquad
    \subfloat[\centering The wheel made weird noises while the mechanic turned.]{{\includegraphics[width=3cm]{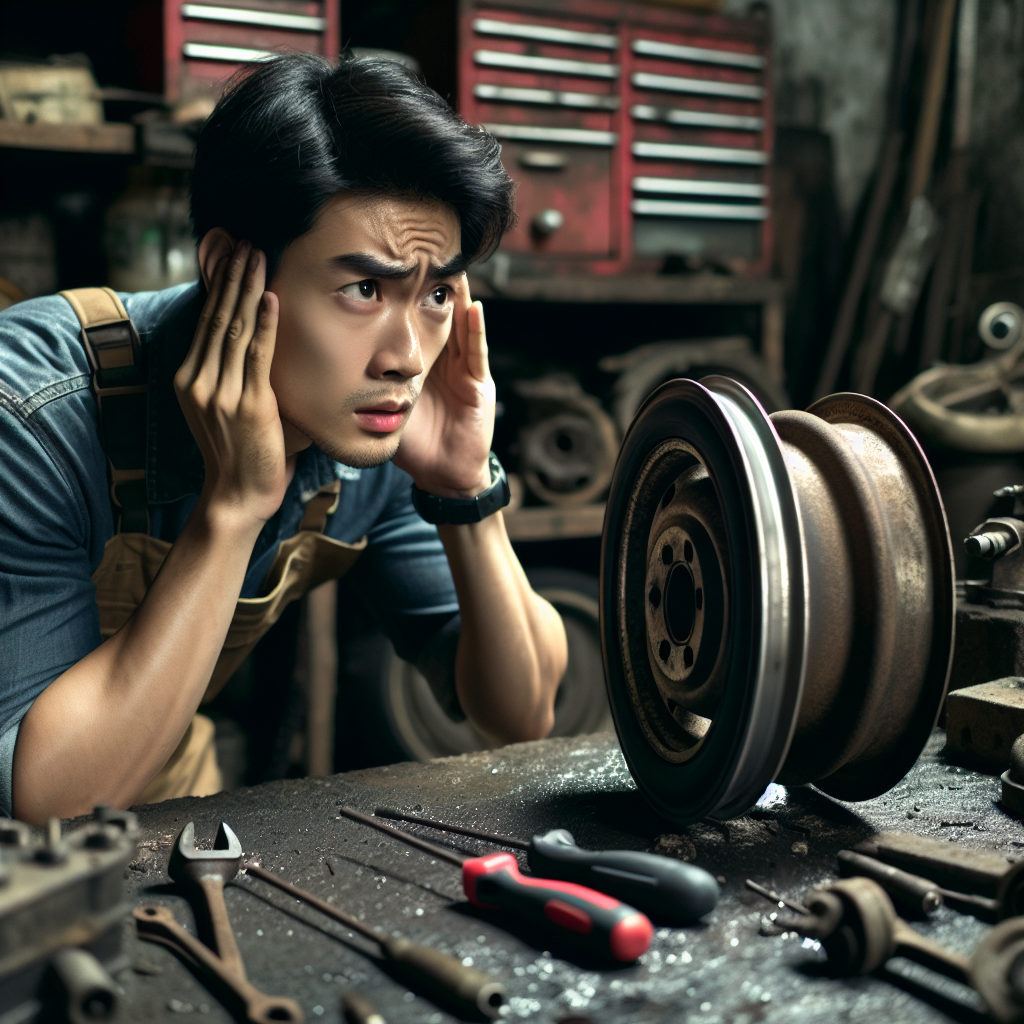} }}
    \caption{Example image for each of our manually-assigned labels for text-to-image generation examples. From left to right: correctly understood, partial misunderstanding, complete misunderstanding and not applicable.}
    \label{fig:dalle_classes}
\end{figure*}

\begin{figure}[t!]
    \centering
    \includegraphics[width=7.0cm, height=5cm, trim={0.4cm 0.2cm 1.6cm 2.5cm}, clip]{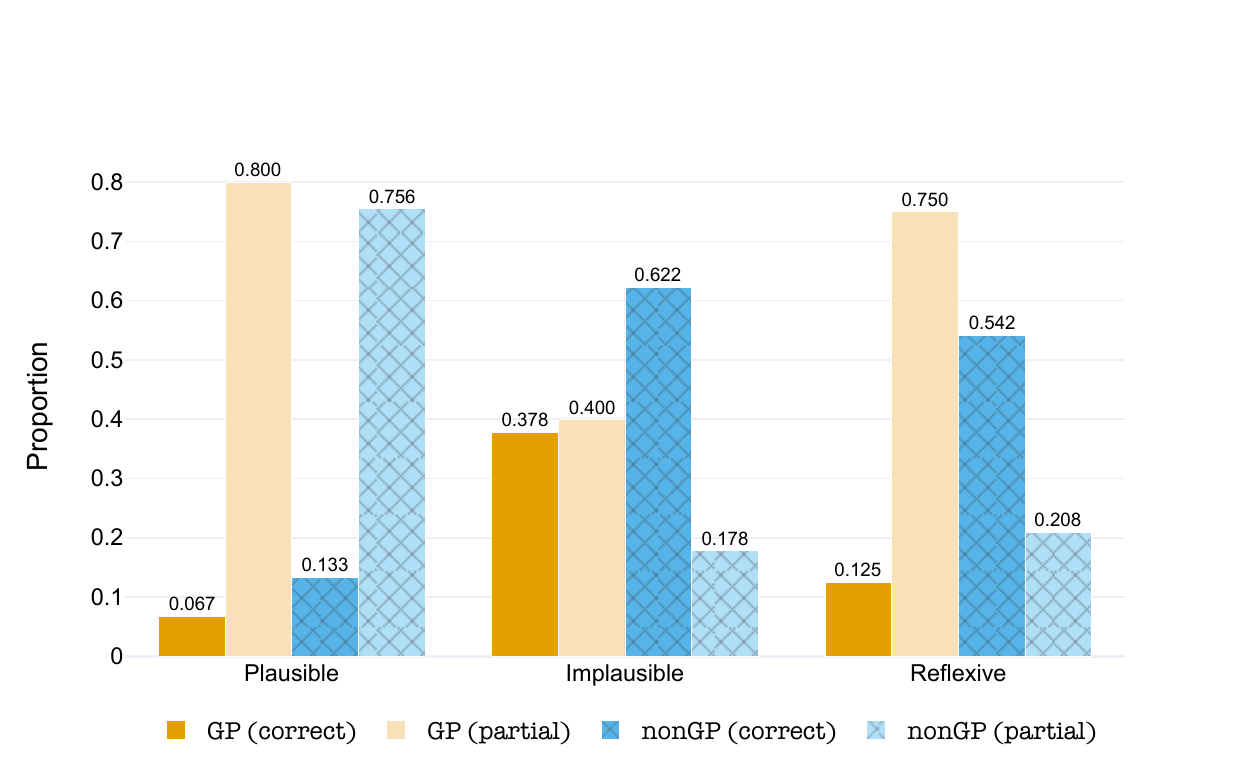}
    \caption{Proportion of images classified as correctly understood and partial understanding for our experimental conditions}
    \label{fig:drawing_classes}
    \vspace{-0.3cm}
\end{figure}

In addition to answering comprehension questions, 
we now test LLM understanding of GP sentences on two additional tasks -- paraphraing and text-to-image generation. Due to cost limitations, we run this experiment on LLMs only. \citet{patson2009lingering} performed a paraphrasing experiment with humans using materials from \citet{christianson2001} and found that paraphrases showed the same misinterpretations that comprehension questions did.

\subsection{Paraphrasing}

We asked LLMs (excluding OLMo-1B and OLMo-7B) to paraphrase a sentence by splitting it into two parts. The correct answer would be to change \emph{``While the man hunted the deer ran into the woods.''} into \emph{``The man hunted. The deer ran into the woods.''}. Using a few-shot prompt, we provided examples of sentences and their breakdowns. The complete prompt is in Appendix \ref{sec:paraphrase_prompt}.

We automatically evaluated the paraphrases using three metrics: a format metric (verifying that the paraphrase consistes of two sentences), a \emph{found-verb} metric (ensuring that the verb \emph{``hunted''} appears in just one sentence), and a \emph{correct-paraphrase} metric (checking that the noun ``deer'' does not appear in the same sentence as ``hunted''). The first two metrics check if the model correctly executed the paraphrasing task, while the last evaluates sentence comprehension. Models that scored below 90\% accuracy in the format and found-verb metrics (Qwen2.5-1.5B, Qwen2.5-3B) were filtered out, and for the remaining  models we measured performance with the correct-paraphrase metric.


Figure \ref{fig:paraphrase_acc} shows the average accuracy across for our model families, categorized by sentence type. We observe that GP sentences are universally more challenging to paraphrase for all models and sentence types. For almost all LLMs, the accuracy on the non-GP sentences is above 90\% across all the manipulations. Moreover, among the GP sentences, implausible ones have the highest paraphrase accuracy across models. Sentences like \emph{``The man hunted the child.''} seem to be too out of distribution for our LLMs to generate.

    We also performed an item-level analysis, where we checked whether the probability assigned to the correct answer for the GP question predicts correct paraphrasing. We measure the AUC for the probability assigned to the correct answer, and find that all our models have an AUC over 0.5, with the minimal AUC being 0.595, the maximal 0.774 and the average 0.696. This suggests that the probability of correctly answering the GP question can strongly predict whether paraphrasing will be accurate. Instruction tuning and model sizes had little effect on the AUC. 

\subsection{Image Generation}

We asked Dall-e-3 \cite{dalle_3}\footnote{We tried 3 other text-to-image models to generate images but the results were not good enough to draw any conclusion.} to generate images for our experimental sentences and manually categorized the resulting images into four groups: \emph{``correctly understood''}, where the image accurately depicts the intended meaning of the sentence, e.g., in ``while the man hunted the deer ran into the woods'', the image would depict a man hunting, and a deer running; \emph{``partial misunderstanding''}, where the image reflects the misinterpretation suggested by the garden path, namely, in the above example, a man hunting a deer, and a deer running; \emph{``complete misunderstanding''}, where the interpretation of the main clause is absent, namely for the example above, the image would not depict a deer running; and \emph{``not applicable''}, where some elements of the sentence are missing, e.g. in the example above there would either be no man, or no deer. Figure \ref{fig:dalle_classes} provides examples for each category. 

Figure \ref{fig:drawing_classes} shows the proportion of the correctly understood and partially understood classes. 
Looking at the partial-misunderstanding label, which corresponds to the type of misinterpretation we have been investigating, we see trends that are similar to the  comprehension questions case. The proportion of examples with this label is lower for non-GP than for GP sentences and for sentences with implausible compared to plausible nouns. In addition, the gap between GP and non-GP sentences is larger for the reflexive/unaccusative case. This further supports the empirical results we saw for comprehension questions and paraphrasing.

\section{Conclusion}

This study explores similarities between human and LLM sentence processing. By focusing on comprehension of garden-path sentences, known for their syntactic complexity and inherent challenges for human processing, we studied whether LLMs have similar difficulties to humans. Our findings demonstrate that humans and LLMs struggle with similar syntactic structures, and notably, some LLMs approximate human behavior quite closely, as indicated by strong correlation metrics. Additionally, the correlation between errors in the comprehension questions,  paraphrasing and image generations tasks suggests shared underlying mechanisms of sentence misinterpretation between the tasks. Our approach not only adds a novel dimension to the evaluation of LLMs but also opens up possibilities for utilizing these models to gain deeper insights into human linguistic processing.
\section*{Limitations}

In our study, we evaluated reading comprehension across a wide array of LLMs. However, the latest state-of-the-art models appeared too late to be incorporated into this version of our paper. Assessing their understanding of our sets would be interesting for a future version of this paper. Additionally, our focus was limited to measuring reading comprehension on Subject/Object GP sentences. Exploring LLMs' comprehension of other types of GP sentences would also be interesting. Finally, we did not collect data on metrics beyond reading comprehension, such as eye gaze or reading time, in our experiments. Gathering such metrics and analyzing their correlation with sentence comprehension could provide valuable insights.

\bibliography{custom}
\bibliographystyle{acl_natbib}
\newpage
\appendix

\section{Transitivity Factors of Verbs}
\label{sec:verb_factors}

In Table \ref{tab:verbTransitivity} we show the transitivity factors of the verbs we use.

\begin{table}[h!]
\centering
\begin{tabular}{llll}

\midrule
lead & 0.314 & start & 0.308 \\ 
leave & 0.673 & drive & 0.426 \\ 
walk & 0.126 & help & 0.256 \\ 
count & 0.371 & perform & 0.411 \\ 
write & 0.336 & smoke & 0.412 \\ 
paint & 0.396 & study & 0.543 \\ 
play & 0.456 & film & 0.235 \\ 
bake & 0.314 & ride & 0.334 \\ 
sail & 0.103 & explore & 0.775 \\ 
fight & 0.334 & attack & 0.626 \\ 
draw & 0.486 & read & 0.502 \\ 
practice & 0.557 & teach & 0.550 \\ 
knit & 0.125 & feed & 0.310 \\ 
sing & 0.404 & order & 0.557 \\ 
check & 0.455 & observe & 0.301 \\ 
sniff & 0.497 & eat & 0.543 \\ 
finish & 0.390 & clean & 0.557 \\ 
type & 0.444 & chew & 0.316 \\ 
hunt & 0.404 & pull & 0.503 \\ 
cheer & 0.378 & photograph & 0.472 \\ 
park & 0.246 & harvest & 0.352 \\ 
drink & 0.497 & cook & 0.260 \\ 
race & 0.169 & stir & 0.666 \\ 
wrestle & 0.334 & steer & 0.612 \\ 
mop & 0.560 & swallow & 0.519 \\ 
dust & 0.255 & vacuum & 0.472 \\ 
whittle & 0.362 & paddle & 0.279 \\ 
juggle & 0.640 \\ 
\bottomrule
\end{tabular}
\caption{Transitivity factors of verbs.}
\label{tab:verbTransitivity}
\end{table}
\newpage
\section{All sentences}
\label{sec:appendix_sents}

In Table \ref{tab:exp_sentences}, we show all the 228 sentences used in our experiments. \\ 

\clearpage 
\onecolumn

\begin{center}
    \scriptsize
    \rowcolors{1}{gray!15}{white} 
    \begin{xtabular*}{\textwidth}{p{0.45\textwidth}|p{0.45\textwidth}}
        \hline
        \textbf{GP structure} & \textbf{Non-GP structure} \\
        \hline
        While the secretary typed the memo neared completion. & The memo neared completion while the secretary typed. \\
        While the secretary typed the drawing neared completion. & The drawing neared completion while the secretary typed. \\
        While the explorer paddled the canoe headed toward a waterfall. & The canoe headed toward a waterfall while the explorer paddled. \\
        While the explorer paddled the yacht headed toward a waterfall. & The yacht headed toward a waterfall while the explorer paddled. \\
        While the cowboy rode the horse sweated profusely. & The horse sweated profusely while the cowboy rode. \\
        While the cowboy rode the leopard sweated profusely. & The leopard sweated profusely while the cowboy rode. \\
        While Tom grilled the hot dog began to burn. & The hot dog began to burn while Tom grilled. \\
        While Tom grilled the barn began to burn. & The barn began to burn while Tom grilled. \\
        While the architect drew the building represented modern times style. & The building represented modern times style while the architect drew. \\
        While the architect drew the melody represented modern times style. & The melody represented modern times style while the architect drew. \\
        While the chef stirred the soup boiled over. & The soup boiled over while the chef stirred. \\
        While the chef stirred the acid boiled over. & The acid boiled over while the chef stirred. \\
        While the tourist explored the tunnel echoed with mysterious sounds. & The tunnel echoed with mysterious sounds while the tourist explored. \\
        While the tourist explored the trunk echoed with mysterious sounds. & The trunk echoed with mysterious sounds while the tourist explored. \\
        While the woman drank the water spilled on the floor. & The water spilled on the floor while the woman drank. \\
        While the woman drank the mud spilled on the floor. & The mud spilled on the floor while the woman drank. \\
        While the snake swallowed the frog kicked vigorously. & The frog kicked vigorously while the snake swallowed. \\
        While the snake swallowed the hippo kicked vigorously. & The hippo kicked vigorously while the snake swallowed. \\
        While the lion attacked the baboon screamed in terror. & The baboon screamed in terror while the lion attacked. \\
        While the lion attacked the penguin screamed in terror. & The penguin screamed in terror while the lion attacked. \\
        While the maid dusted the picture tipped over. & The picture tipped over while the maid dusted. \\
        While the maid dusted the cow tipped over. & The cow tipped over while the maid dusted. \\
        While the teacher counted the children formed a line. & The children formed a line while the teacher counted. \\
        While the teacher counted the ants formed a line. & The ants formed a line while the teacher counted. \\
        While the champion raced the challenger stumbled and fell. & The challenger stumbled and fell while the champion raced. \\
        While the champion raced the spectator stumbled and fell. & The spectator stumbled and fell while the champion raced. \\
        While Jerry played the violin went out of tune. & The violin went out of tune while Jerry played. \\
        While Jerry played the speakers went out of tune. & The speakers went out of tune while Jerry played. \\
        While the girl painted the rainbow slowly faded outside. & The rainbow slowly faded outside while the girl painted. \\
        While the girl painted the music slowly faded outside. & The music slowly faded outside while the girl painted. \\
        While Kendra parked the van bumped the curb. & The van bumped the curb while Kendra parked. \\
        While Kendra parked the horse bumped the curb. & The horse bumped the curb while Kendra parked. \\
        While Angela cleaned the dog stood in the yard. & The dog stood in the yard while Angela cleaned. \\
        While Angela cleaned the mailman stood in the yard. & The mailman stood in the yard while Angela cleaned. \\
        While the sailor smoked the pipe glowed brightly. & The pipe glowed brightly while the sailor smoked. \\
        While the sailor smoked the firefly glowed brightly. & The firefly glowed brightly while the sailor smoked. \\
        While the tourist filmed the dancer blocked the sidewalk. & The dancer blocked the sidewalk while the tourist filmed. \\
        While the tourist filmed the trashbags blocked the sidewalk. & The trashbags blocked the sidewalk while the tourist filmed. \\
        While the athlete wrestled the opponent shouted insults. & The opponent shouted insults while the athlete wrestled. \\
        While the athlete wrestled the audience shouted insults. & The audience shouted insults while the athlete wrestled. \\
        While the warrior fought the enemy retreated. & The enemy retreated while the warrior fought. \\
        While the warrior fought the rhino retreated. & The rhino retreated while the warrior fought. \\
        While Harry chewed the steak fell to the floor. & The steak fell to the floor while Harry chewed. \\
        While Harry chewed the book fell to the floor. & The book fell to the floor while Harry chewed. \\
        While Bill ate the turkey sat on the table. & The turkey sat on the table while Bill ate. \\
        While Bill ate the baby sat on the table. & The baby sat on the table while Bill ate. \\
        While Janet baked the bread rose in the oven. & The bread rose in the oven while Janet baked. \\
        While Janet baked the pressure rose in the oven. & The pressure rose in the oven while Janet baked. \\
        While the public cheered the team left the restaurant. & The team left the restaurant while the public cheered. \\
        While the public cheered the rats left the restaurant. & The rats left the restaurant while the public cheered. \\
        While the cleaner mopped the floor was filled with stains. & The floor was filled with stains while the cleaner mopped. \\
        While the cleaner mopped the ceiling was filled with stains. & The ceiling was filled with stains while the cleaner mopped. \\
        While the chef cooked the meal impressed the couple. & The meal impressed the couple while the chef cooked. \\
        While the chef cooked the cat impressed the couple. & The cat impressed the couple while the chef cooked. \\
        While the child finished the homework waited on the table. & The homework waited on the table while the child finished. \\
        While the child finished the tax forms waited on the table. & The tax forms waited on the table while the child finished. \\
        While the student knitted the sweater sold to the highest bidder. & The sweater sold to the highest bidder while the student knitted. \\
        While the student knitted the painting sold to the highest bidder. & The painting sold to the highest bidder while the student knitted. \\
        While the astronomer observed the comet lit up the room. & The comet lit up the room while the astronomer observed. \\
        While the astronomer observed the lightbulb lit up the room. & The lightbulb lit up the room while the astronomer observed. \\
        While the players started the game bored the children. & The game bored the children while the players started. \\
        While the players started the lesson bored the children. & The lesson bored the children while the players started. \\
        While the professor taught the students looked at the board. & The students looked at the board while the professor taught. \\
        While the professor taught the puppies looked at the board. & The puppies looked at the board while the professor taught. \\
        While the pianist practiced the melody echoed through the hall. & The melody echoed through the hall while the pianist practiced. \\
        While the pianist practiced the vacuum echoed through the hall. & The vacuum echoed through the hall while the pianist practiced. \\
        While the couple left the bar buzzed with activity. & The bar buzzed with activity while the couple left. \\
        While the couple left the hive buzzed with activity. & The hive buzzed with activity while the couple left. \\
        While the gardener harvested the tomatoes hanged on the vine. & The tomatoes hanged on the vine while the gardener harvested. \\
        While the gardener harvested the chrysalis hanged on the vine. & The chrysalis hanged on the vine while the gardener harvested. \\
        While the horse pulled the cart moved silently. & The cart moved silently while the horse pulled. \\
        While the horse pulled the submarine moved silently. & The submarine moved silently while the horse pulled. \\
        While the man hunted the deer ran into the woods. & The deer ran into the woods while the man hunted. \\
        While the man hunted the child ran into the woods. & The child ran into the woods while the man hunted. \\
        While the skipper sailed the boat veered off course. & The boat veered off course while the skipper sailed. \\
        While the skipper sailed the car veered off course. & The car veered off course while the skipper sailed. \\
        While the orchestra performed the symphony played on the radio. & The symphony played on the radio while the orchestra performed. \\
        While the orchestra performed the newsflash played on the radio. & The newsflash played on the radio while the orchestra performed. \\
        While the bridesmaid ordered the dress got delivered. & The dress got delivered while the bridesmaid ordered. \\
        While the bridesmaid ordered the tractor got delivered. & The tractor got delivered while the bridesmaid ordered. \\
        While Susan wrote the letter fell off the table. & The letter fell off the table while Susan wrote. \\
        While Susan wrote the Bible fell off the table. & The Bible fell off the table while Susan wrote. \\
        While the farmer steered the tractor pulled the car. & The tractor pulled the car while the farmer steered. \\
        While the farmer steered the tank pulled the car. & The tank pulled the car while the farmer steered. \\
        While the lawyer studied the contract lay on the roll-top desk. & The contract lay on the roll-top desk while the lawyer studied. \\
        While the lawyer studied the cat lay on the roll-top desk. & The cat lay on the roll-top desk while the lawyer studied. \\
        While the clown juggled the balls fell on the ground. & The balls fell on the ground while the clown juggled. \\
        While the clown juggled the cats fell on the ground. & The cats fell on the ground while the clown juggled. \\
        While Anne vacuumed the rug lost its colors. & The rug lost its colors while Anne vacuumed. \\
        While Anne vacuumed the chameleon lost its colors. & The chameleon lost its colors while Anne vacuumed. \\
        While Jim bathed the child giggled with delight. & The child giggled with delight while Jim bathed. \\
        While the chimps groomed the baboons sat in the grass. & The baboons sat in the grass while the chimps groomed. \\
        While Frank dried off the car sat in the driveway. & The car sat in the driveway while Frank dried off. \\
        While Betty woke up the neighbor coughed loudly. & The neighbor coughed loudly while Betty woke up. \\
        While the thief hid the jewelry sparkled brightly. & The jewelry sparkled brightly while the thief hid. \\
        While Anna dressed the baby spit up on the bed. & The baby spit up on the bed while Anna dressed. \\
        While the boy washed the dog barked loudly. & The dog barked loudly while the boy washed. \\
        While the jockey settled down the horse stood in the stall. & The horse stood in the stall while the jockey settled down. \\
        While the mother undressed the baby cried softly. & The baby cried softly while the mother undressed. \\
        While the nurse shaved the patient watched TV. & The patient watched TV while the nurse shaved. \\
        While the girl scratched the cat stared at the dog. & The cat stared at the dog while the girl scratched. \\
        While the mother calmed down the children sat on the bed. & The children sat on the bed while the mother calmed down. \\
        While the artist changed the painting stood the test of time. & The painting stood the test of time while the artist changed. \\
        While the child grew the tomatoes invaded the garden. & The tomatoes invaded the garden while the child grew. \\
        While the toddler rolled the ball kept moving across the room. & The ball kept moving across the room while the toddler rolled. \\
        While the mechanic turned the wheel made weird noises. & The wheel made weird noises while the mechanic turned. \\
        While the ice cap shrank the river stayed warm. & The river stayed warm while the ice cap shrank. \\
        While the gymnast swung the bat fell on the road. & The bat fell on the road while the gymnast swung. \\
        While Bill trained the athletes started the game. & The athletes started the game while Bill trained. \\
        While the dog moved the bone went bad. & The bone went bad while the dog moved. \\
        While the river flooded the valley kept on living quietly. & The valley kept on living quietly while the river flooded. \\
        While the log broke the window moved on its axis. & The window moved on its axis while the log broke. \\
        While the match lit the fire warmed the children. & The fire warmed the children while the match lit. \\
        While the wave crashed the boat was stranded in the ocean. & The boat was stranded in the ocean while the wave crashed. \\
    \label{tab:exp_sentences}
    \end{xtabular*}
\end{center}

\twocolumn
\newpage
\section{Generalized Linear Mixed-Effect Models}
\label{sec:glmm_app}

In Table \ref{tab:glmm} we show the different Generalized Linear Mixed-Effects Models we performed. The different variables are: SentenceType (GP vs. non-GP), ManipulationType (plausible, implausible, reflexive), TransitiveFactor (the transitive factor of the verb in the sentence).

\begin{table*}[t!]
    \scriptsize 
    \centering
        \begin{tabular}{ l | l | l }
         \textbf{Data} & \textbf{Formula} & \textbf{Effect} \\
        \thickhline
        Plausible & SentenceType + (1 + SentenceType | SetId) & SentenceType: Estimate -0.46, P-Value: 0.069 \\
        Implausible & SentenceType + (1 + SentenceType | SetId) & SentenceType: Estimate -0.43, P-Value: 0.019 \\
        Reflexive & SentenceType + (1 + SetId) & SentenceType: Estimate -1.51, P-Value: 2.58e-13 \\
        Plausible and Implausible & ManipulationType + (1 + ManipulationType | SetId) & ManipulationType: Estimate -2.13, P-Value: 4.11e-16 \\
        Plausible and Reflexive - GP & ManipulationType + (1 + ManipulationType | SetId) & ManipulationType: Estimate 1.40, P-Value: 1.35e-5 \\
        Plausible and Reflexive - nonGP & ManipulationType + (1 + ManipulationType | SetId) & ManipulationType: Estimate 2.41, P-Value: 2.45e-14 \\
        \end{tabular}
        \vspace{-0.2cm}
    \caption{Details about the Generalized Linear Mixed-Effects Models we performed}
    \label{tab:glmm}
\end{table*}

\section{Prompt example}
\label{sec:prompt_example}

Figure \ref{fig:prompt} shows an example prompt.

\begin{figure*}
\colorbox{gray!10}{
\begin{minipage}{15cm}
\texttt{You are a linguistic experiment subject. You will be presented with a sentence, and will need to answer a reading comprehension question. You will need to select an option amongst the proposed answers. \\
Here are a few examples of questions and relevant answers: \\ \\
The doctor that the nurse called checked on the patient yesterday. \\
Answer with Yes or No: \\
Did the nurse call the doctor? \\
Yes \\ \\
The teacher that helped the student graded the papers on the weekend. \\
Answer with Yes or No: \\
Did the student grade the papers? \\
No \\ \\
The sailor that the captain punished stayed in his room. \\ 
Answer with Yes or No: \\
Did the captain stay in his room? \\
No \\ \\
The driver that saved the cyclist went back home. \\
Answer with Yes or No: \\
Did the driver go back home? \\
Yes \\ \\
Here is the sentence: \\
While the secretary typed the memo neared completion. \\ \\
Answer this question: \\
Did the memo near completion? \\ \\
My answer is: 
}
\end{minipage}}
\caption{Example of a prompt}
\label{fig:prompt}
\end{figure*}
\section{Model results}
\label{sec:all_model_res}

In Table \ref{tab:all_model_res} we show the results of all the models on our conditions.

\begin{table*}[t!]
    \scriptsize 
    \centering
        \begin{tabular}{ l | c c c c c c }
         \textbf{Model} & \textbf{Plausible - GP} & \textbf{Plausible - nonGP} & \textbf{Implausible - GP} & \textbf{Implausible - nonGP} & \textbf{Reflexive - GP} & \textbf{Reflexive - nonGP} \\
        \thickhline
GPT4 & 0.087 & 0.157 & 0.620 & 0.827 & 0.223 & 0.825 \\ 
GPT4-T & 0.327 & 0.472 & 0.849 & 0.958 & 0.679 & 0.940 \\ 
GPT4-O & 0.172 & 0.121 & 0.792 & 0.833 & 0.384 & 0.708 \\ 
GPT4-O-M & 0.058 & 0.418 & 0.529 & 0.928 & 0.231 & 1.000 \\ 
O1-Mini & 0.138 & 0.342 & 0.573 & 0.880 & 0.442 & 0.933 \\ 
O1-Prev. & 0.680 & 0.436 & 0.889 & 0.916 & 0.917 & 0.992 \\ \hline
Qwen-1.5B & 0.167 & 0.239 & 0.291 & 0.492 & 0.224 & 0.388 \\ 
Qwen-1.5B-Inst & 0.000 & 0.069 & 0.096 & 0.572 & 0.001 & 0.350 \\ 
Qwen-3B & 0.231 & 0.268 & 0.403 & 0.703 & 0.284 & 0.490 \\ 
Qwen-3B-Inst & 0.026 & 0.279 & 0.341 & 0.913 & 0.107 & 0.740 \\
Qwen-7B & 0.317 & 0.321 & 0.592 & 0.731 & 0.426 & 0.686 \\ 
Qwen-7B-Inst & 0.136 & 0.338 & 0.576 & 0.855 & 0.354 & 0.820 \\ 
Qwen-14B & 0.265 & 0.220 & 0.635 & 0.744 & 0.359 & 0.605 \\ 
Qwen-14B-Inst & 0.231 & 0.197 & 0.687 & 0.835 & 0.380 & 0.792 \\ 
Qwen-32B & 0.263 & 0.282 & 0.751 & 0.825 & 0.452 & 0.727 \\ 
Qwen-32B-Inst & 0.358 & 0.426 & 0.848 & 0.933 & 0.607 & 0.924 \\  
Qwen-72B & 0.290 & 0.340 & 0.662 & 0.803 & 0.456 & 0.677 \\ 
Qwen-72B-Inst & 0.229 & 0.334 & 0.807 & 0.918 & 0.505 & 0.873 \\ \hline
Gemma-2B & 0.068 & 0.080 & 0.087 & 0.120 & 0.070 & 0.097 \\ 
Gemma-2B-Inst & 0.002 & 0.058 & 0.054 & 0.521 & 0.001 & 0.276 \\ 
Gemma-9B & 0.149 & 0.224 & 0.364 & 0.550 & 0.203 & 0.462 \\ 
Gemma-9B-Inst & 0.022 & 0.132 & 0.269 & 0.791 & 0.038 & 0.772 \\ 
Gemma-27B & 0.570 & 0.570 & 0.845 & 0.921 & 0.712 & 0.913 \\ 
Gemma-27B-Inst & 0.060 & 0.167 & 0.421 & 0.831 & 0.123 & 0.901 \\ \hline
Llama3.2-1B & 0.247 & 0.295 & 0.269 & 0.329 & 0.270 & 0.317 \\ 
Llama3.2-1B-Inst & 0.140 & 0.201 & 0.237 & 0.360 & 0.183 & 0.255 \\ 
Llama3.2-3B & 0.247 & 0.304 & 0.331 & 0.426 & 0.296 & 0.392 \\ 
Llama3.2-3B-Inst & 0.243 & 0.430 & 0.463 & 0.798 & 0.332 & 0.776 \\ 
Llama3.2-11B-Vis & 0.519 & 0.587 & 0.720 & 0.870 & 0.618 & 0.823 \\ 
Llama3.2-11B-Vis-Inst & 0.531 & 0.618 & 0.797 & 0.918 & 0.676 & 0.895 \\ 
Llama3.2-90B-Vis & 0.419 & 0.288 & 0.775 & 0.871 & 0.654 & 0.847 \\ 
Llama3.2-90B-Vis-Inst & 0.378 & 0.242 & 0.790 & 0.892 & 0.693 & 0.883 \\ \hline
Olmo-7B-Tokens-8B & 0.655 & 0.665 & 0.654 & 0.663 & 0.649 & 0.657 \\
Olmo-7B-Tokens-111B & 0.074 & 0.076 & 0.076 & 0.078 & 0.076 & 0.076 \\ 
Olmo-7B-Tokens-446B & 0.225 & 0.254 & 0.237 & 0.277 & 0.229 & 0.268 \\ 
Olmo-7B-Tokens-599B & 0.159 & 0.175 & 0.179 & 0.211 & 0.167 & 0.193 \\ 
Olmo-7B-Tokens-1384B & 0.043 & 0.056 & 0.059 & 0.100 & 0.053 & 0.080 \\ 
Olmo-7B-Tokens-1656B & 0.102 & 0.129 & 0.166 & 0.249 & 0.129 & 0.226 \\ 
Olmo-7B-Tokens-1920B & 0.179 & 0.190 & 0.227 & 0.288 & 0.205 & 0.238 \\ 
Olmo-7B-Tokens-2176B & 0.246 & 0.272 & 0.282 & 0.368 & 0.260 & 0.308 \\ 
Olmo-7B-Tokens-2716B & 0.139 & 0.164 & 0.203 & 0.326 & 0.163 & 0.239 \\ 
Olmo-7B-Tokens-2729B & 0.122 & 0.142 & 0.171 & 0.265 & 0.141 & 0.204 \\ \hline
Olmo-1B-Tokens-41B & 0.153 & 0.160 & 0.155 & 0.162 & 0.152 & 0.160 \\ 
Olmo-1B-Tokens-794B & 0.140 & 0.142 & 0.148 & 0.156 & 0.142 & 0.148 \\ 
Olmo-1B-Tokens-1169B & 0.204 & 0.230 & 0.209 & 0.242 & 0.201 & 0.237 \\ 
Olmo-1B-Tokens-1547B & 0.248 & 0.267 & 0.253 & 0.283 & 0.251 & 0.283 \\ 
Olmo-1B-Tokens-1922B & 0.280 & 0.293 & 0.285 & 0.300 & 0.287 & 0.301 \\ 
Olmo-1B-Tokens-2176B & 0.295 & 0.328 & 0.303 & 0.348 & 0.306 & 0.348 \\ 
Olmo-1B-Tokens-2364B & 0.280 & 0.313 & 0.286 & 0.326 & 0.282 & 0.320 \\ 
Olmo-1B-Tokens-2742B & 0.166 & 0.198 & 0.174 & 0.221 & 0.172 & 0.219 \\ 
Olmo-1B-Tokens-2932B & 0.130 & 0.155 & 0.134 & 0.168 & 0.129 & 0.158 \\ 
Olmo-1B-Tokens-3048B & 0.452 & 0.482 & 0.471 & 0.515 & 0.470 & 0.499 \\ 
        \end{tabular}
        \vspace{-0.2cm}
    \caption{Results of the models on our manipulations}
    \label{tab:all_model_res}
\end{table*}

\section{Paraphrase prompt example}
\label{sec:paraphrase_prompt}

Figure \ref{fig:paraphrase_prompt} shows an example prompt for the paraphrasing task.

\begin{figure*}
\colorbox{gray!10}{
\begin{minipage}{15cm}
\texttt{You are a linguistic experiment subject. You will be presented with a sentence, and you will need to split it into two sentences that convey the exact same situation as the original sentence. You will be provided with a few examples. Note: keep the sentences as simple as possible. \\ \\
Example 1: \\
Sentence: The dog ran in the courtyard, and the man fell. \\
Splitted: \\
1. The dog ran in the courtyard. \\
2. The man fell.\\ \\
Example 2: \\
Sentence: Sarah met her boss in the park when the plane crashed. \\
Splitted:  \\
1. Sarah met her boss in the park. \\ 
2. The plane crashed. \\ \\
Example 3: \\
Sentence: She cleaned the mess that her sister made. \\
Splitted: \\
1. She cleaned the mess. \\
2. Her sister made the mess. \\ \\
Example 4: \\
Sentence: They looked for the treasure, hoping to find salvation. \\
Splitted: \\
1. The looked for the treasure. \\
2. They hoped to find salvation.
}
\end{minipage}}
\caption{Example of a paraphrase task prompt}
\label{fig:paraphrase_prompt}
\end{figure*}

\end{document}